\documentclass{article}
\usepackage{ijcai20}
\usepackage{times}
\usepackage{soul}
\usepackage{url}
\usepackage[hidelinks]{hyperref}
\usepackage[utf8]{inputenc}
\usepackage[small]{caption}
\usepackage{graphicx}
\usepackage{amsmath}
\usepackage{amsthm,amssymb}
\usepackage{amsfonts}
\usepackage{booktabs}
\usepackage{comment}
\usepackage{color}
\usepackage{makecell}
\newcommand{\hide}[1]{}

\usepackage{algpseudocode,algorithm,algorithmicx}
\algrenewcommand\algorithmicrequire{\textbf{Input:}}

\DeclareMathOperator*{\argmin}{argmin}

\title{Task-Based Learning via Task-Oriented Prediction Network\\ with Applications in Finance}

\author{
Di Chen\footnote{The work was done during the first author's internship at IBM Research. The views and conclusions are those of the authors and should not be interpreted as that of IBM Research.}$^1$\and
Yada Zhu$^2$\and
Xiaodong Cui$^2$\And
Carla P. Gomes$^1$
\affiliations
$^1$Cornell University\\
$^2$IBM T. J. Watson Research Center
\emails
di@cs.cornell.edu,
\{yzhu, cuix\}@us.ibm.com,
gomes@cs.cornell.edu
}

\begin{document}

\maketitle

\begin{abstract}
Real-world applications often involve domain-specific and task-based performance objectives that are not captured by the standard machine learning losses, but are critical for decision making. A key challenge for direct integration of more meaningful domain and task-based evaluation criteria into an end-to-end gradient-based training process is the fact that often such performance objectives are not necessarily differentiable and may even require additional decision-making optimization processing. We propose the Task-Oriented Prediction Network (TOPNet), an end-to-end learning scheme that automatically integrates task-based evaluation criteria into the learning process via a learnable surrogate loss function, which directly guides the model towards the task-based goal. A major benefit of the proposed TOPNet learning scheme lies in its capability of automatically integrating non-differentiable evaluation criteria, which makes it particularly suitable for diversified and customized task-based evaluation criteria in real-world tasks. We validate the performance of TOPNet on two real-world financial prediction tasks, revenue surprise forecasting and credit risk modeling. The experimental results demonstrate that TOPNet significantly outperforms both traditional modeling with standard losses and modeling with hand-crafted heuristic differentiable surrogate losses.

\end{abstract}

\section{Introduction}



Prediction models have been widely used to facilitate decision making across domains, e.g., retail demand prediction for inventory control~\cite{Riemer:2016:CFM:3045390.3045707}, user behavior prediction for display advertisement~\cite{Yang:2017:LAU:3097983.3098089}, and financial market movement prediction for portfolio management~\cite{marcos2018}, to name a few.
These models are often trained using standard machine learning loss functions, such as mean square error (MSE), mean absolute error (MAE) and cross-entropy loss (CE).
However, these criteria commonly used to train prediction models are often different from the task-based criteria used to evaluate model performance \cite{bengio1997using,donti2017task}.
For instance, a standalone image classification model is often trained by optimizing cross-entropy loss. 
However, when it is used to guide autonomous driving, we may care more about misclassifying a traffic sign vs. misclassifying a garbage can.
In revenue surprise forecasting, financial institutes often train a regression model to predict the revenue surprise for each public company minimizing mean square error. 
However, they evaluate the model performance based on the \textit{Directional Accuracy} 
(percentage of predictions that are more directional accurate) 
and the \textit{Magnitude Accuracy} 
(percentage of predictions that are 50\% more accurate)
with respect to industry benchmarks (e.g. the consensus of the Wall Street analysts\footnote{https://www.investopedia.com/terms/c/consensusestimate.asp}), which provide more value for downstream portfolio management.
In loan default risk modeling, banks often train a classification model to predict the default probability of each loan application, and optimize the probability threshold to accept/reject loans with low/high risk.
Eventually, they evaluate the model performance by aggregating the total profit made from those loans.

Despite the popularity of standard machine learning losses, models trained with such standard losses are not necessarily aligned with the task-based evaluation criteria and as a result may perform poorly with respect to the ultimate task-based objective.
One straightforward solution to this problem is to directly use the task-based evaluation criteria as the loss function.
However, task-based evaluation criteria are often unfriendly to an end-to-end gradient-based training process due to the fact that often such performance objectives are not necessarily differentiable and may even require additional decision-making optimization processing.
Existing works \cite{elmachtoub2017smart,bengio1997using,donti2017task,wilder2019melding,perrault2019decision,wilder2019end} in this area mainly focus on deriving heuristic surrogate loss functions that differentiate from downstream evaluation criteria to the upstream prediction model via certain relaxations or KKT conditions.
However, those derivations are mainly hand-crafted and task-specific.
As a result, it requires a considerable amount of effort to find proper surrogate losses for new tasks, especially when the evaluation criteria are complicated or involve non-convex optimization.
Moreover, hand-crafted surrogate losses  are not optimized, which can hardly become the optimal choice.
Therefore, a general end-to-end learning scheme, which can automatically integrate the task-based evaluation criteria, is still needed.

\textit{\textbf{ 
We propose the Task-Oriented Prediction Network (TOPNet), a generic end-to-end learning scheme that  automatically integrates task-based evaluation criteria into the learning process via a learnable differentiable surrogate loss function, which approximates the true task-based loss and directly guides the prediction model to the task-based goal.}}
Specifically, {\textbf{(i)}} TOPNet learns a differentiable surrogate loss function
%
parameterized by a task-oriented loss estimator network that 
approximates the true task-based loss
given the prediction, the ground-truth label and necessary contextual information.
{\textbf{(ii)}} TOPNet optimizes a predictor using the learned surrogate loss function, to approximately optimize its performance w.r.t.\ the true task-based loss.
{\textbf{(iii)}} We demonstrate the performance of TOPNet on two real-world financial prediction tasks: a revenue surprise forecasting task and a credit risk modeling task, where the former is a regression task and the latter is a classification task.
Applying TOPNet to these two tasks, we show that TOPNet significantly boosts the ultimate task-based goal by integrating the task-based evaluation criteria, outperforming both traditional modeling with standard losses and modeling with heuristic differentiable (relaxed) surrogate losses.


\section{Related Work}

Integrating task-based evaluation criteria into the learning process was studied under different names, such as \textit{task-based learning} and \textit{decision-focused learning}.
The earliest work \cite{bengio1997using}, which is closely related to ours, optimizes the neural network based on returns obtained via a hedging strategy, to predict financial prices.
Later, \citeauthor{kao2009directed} (\citeyear{kao2009directed}) proposed Directed Regression, which minimizes a convex combination of least square loss and a task-based loss, 
to achieve a better regression performance w.r.t.\ the decision objective.
%
\citeauthor{elmachtoub2017smart} (\citeyear{elmachtoub2017smart}) derived a convex surrogate loss function called SPO+ loss via duality theory, 
to leverage the upstream prediction model and the downstream optimization task for linear programming.
%
\citeauthor{donti2017task} (\citeyear{donti2017task}) proposed task-based model learning for stochastic programming, where they differentiate through the KKT condition of the convex objective, to provide gradients for the upstream prediction model to capture the downstream optimization objective.
%
Recent works \cite{perrault2019decision,wilder2019melding,wilder2019end} applied a similar idea to security games, combinatorial optimization problems and graph optimization problems, to integrate the downstream objectives into the upstream modeling.

Those previous works \cite{bengio1997using,elmachtoub2017smart,donti2017task,perrault2019decision,wilder2019melding,wilder2019end} mainly focus on deriving a differentiable surrogate loss function for the downstream evaluation criteria to provide gradients to the upstream prediction model. 
Even though those works have developed many surrogate losses for different evaluation criteria,
their approaches either require the objective to be convex or use hand-crafted relaxation to approximate the ultimate objective.
In contrast, Task-Oriented Prediction Network (TOPNet) does not require hand-crafted differentiation of the downstream evaluation criteria.
Instead, TOPNet learns a differentiable surrogate loss via a task-oriented loss estimator network, which automatically approximates the true task-based loss and directly guides the upstream predictor towards the downstream task-based goal. 
In the context of task-based learning, TOPNet is the first work that automatically integrates the true task-based evaluation criteria into an end-to-end learning process via a learnable surrogate loss function.

\section{Problem Formulation}

We first formally define the task-based prediction problem that we address in this paper. 
We use $\mathbf{x} \in \mathcal{X}\subseteq\mathbb{R}^d$ and $y \in \mathcal{Y}$ for the feature and label variables. 
%
%
Given dataset $D=\{(\mathbf{x}_1, y_1)$, $(\mathbf{x}_2, y_2)$ ...,$(\mathbf{x}_n, y_n)\}$, which is sampled from an unknown data distribution $P$ with density function $p(x,y)$, our prediction task can be formulated as learning a conditional distribution $q_{\theta}(\hat{y}|\mathbf{x})$ that minimizes the expected task-based loss (task-based criteria) 
$\ell^T(q_{\theta}(\hat{y}|\mathbf{x}), p(y|\mathbf{x}), c)$, i.e., 
\begin{equation}
\min\limits_{\theta}
\mathbb{E}_{\mathbf{x}\sim p(\mathbf{x})}
[\ell^T(q_{\theta}(\hat{y}|\mathbf{x}), p(y|\mathbf{x}), c)],
\label{eqn:tbl}
\end{equation}
where $c$ denotes some necessary contextual information related to task-based criteria, $p(\mathbf{x})$ denotes the marginal distribution of $\mathbf{x}$, and $\theta$ denotes the parameters of our prediction model. 
As implied in formulation (\ref{eqn:tbl}), we mainly consider the tasks whose task-based losses can be computed point-wisely.

A key challenge of task-based learning comes from the fact that the true task-based loss function $\ell^T(q_{\theta}(\hat{y}|\mathbf{x}), p(y|\mathbf{x}), c)$ is often non-differentiable and may even involve additional decision-making optimization processing, which cannot be used directly in popular gradient-based learning methods.
For instance, in revenue surprise forecasting, the task-based criteria evaluate a prediction $\hat{y}$ based on both the true revenue surprise $y$ and the prediction of the consensus of the Wall Street analysts $c$ (in that case, both $q_{\theta}(\hat{y}|\mathbf{x})$ and $p(y|\mathbf{x})$ are Dirac delta distribution). 
Specifically, the criteria compute whether the prediction is more directional accurate and whether the prediction is significantly (50\%) more accurate compared with the Wall Street consensus, which both involve non-differentiable functions (see detailed formula in our experiments).
Likewise, in credit risk modeling, the task-based criteria involve optimizing a probability decision threshold $p_D$ to maximize the profit after approving all loan applications with a predicted default probability $p_i$ lower than $p_D$.

A straightforward solution to this challenge is to use a surrogate loss function $\ell^S(q_{\theta}(\hat{y}|\mathbf{x}), p(y|\mathbf{x}), c)$ to replace the true task-based loss and guide the learning process.
Existing works mainly focus on using standard machine learning loss functions, such as mean square error (MSE), mean absolute error (MAE) and cross-entropy loss (CE), or other task-specific differentiable loss functions \cite{bengio1997using,elmachtoub2017smart,donti2017task,perrault2019decision,wilder2019melding,wilder2019end} as the surrogate loss, that is, 
\begin{equation}
\min\limits_{\theta}
\mathbb{E}_{\mathbf{x}\sim p(\mathbf{x})}
[\ell^S(q_{\theta}(\hat{y}|\mathbf{x}), p(y|\mathbf{x}), c)].
\end{equation}
However, both standard machine learning losses and task-specific differentiable losses are selected
manually.
Thus, finding a proper surrogate loss function requires a considerable amount of effort, especially when the evaluation criteria are complicated or involve non-convex optimization.
Therefore, such approaches require considerable customization and do not provide a general methodology to task-based learning. 

\section{Task-Based Learning via A Learnable Differentiable Surrogate Loss}

Instead of manually designing a hand-crafted differentiable loss, we propose to learn a differentiable surrogate loss function $\ell^S_\omega(q_{\theta}(\hat{y}|\mathbf{x}), p(y|\mathbf{x}), c)$ via a neural network parameterized by $\omega$, to approximate the true task-based loss and guide the prediction model. 
Specifically, we formulate the task-based learning problem as a bilevel optimization, i.e., 
\begin{align}
&\min\limits_{\theta}
\mathbb{E}_{\mathbf{x}\sim p(\mathbf{x})}
[\ell^S_{\omega^{*}}(q_{\theta}(\hat{y}|\mathbf{x}), p(y|\mathbf{x}), c)] \label{eqn:upper}
\\
&\mbox{subject to: } \nonumber\\
&
\resizebox{0.9\hsize}{!}{$
\omega^{*}=\argmin\limits_{\omega}
\mathbb{E}_{\mathbf{x}\sim p(\mathbf{x})}
[D(\ell^S_\omega(q_{\theta}(\hat{y}|\mathbf{x}), p(y|\mathbf{x}), c)
||
\ell^T(q_{\theta}(\hat{y}|\mathbf{x}), p(y|\mathbf{x}), c)
)]
$}
\label{eqn:lower}\\
&\mbox{, where $D(\cdot||\cdot)$ is a discrepancy function.
} 
\nonumber
\end{align}
%
In this paper, we assume that both $\ell^S_\omega(q_{\theta}(\hat{y}|\mathbf{x}), p(y|\mathbf{x}), c)$ and $\ell^T(q_{\theta}(\hat{y}|\mathbf{x}), p(y|\mathbf{x}), c)$ are real-valued loss functions. 
Thus, we mainly consider using absolute error loss or square error loss as the discrepancy function, i.e., $D(x||y)=|x-y|$ or $D(x||y)=(x-y)^2$.
\small
\begin{align}
&\mathbb{E}_{\mathbf{x}\sim p(\mathbf{x})}
[\ell^T(q_{\theta}(\hat{y}|\mathbf{x}), p(y|\mathbf{x}), c)] \nonumber\\
\leq 
&\mathbb{E}_{\mathbf{x}\sim p(\mathbf{x})}
[\ell^S_\omega(q_{\theta}(\hat{y}|\mathbf{x}), p(y|\mathbf{x}), c)]
] + \label{eqn:bound1}\\
&\mathbb{E}_{\mathbf{x}\sim p(\mathbf{x})}
[
|\ell^S_\omega(q_{\theta}(\hat{y}|\mathbf{x}), p(y|\mathbf{x}), c) -
\ell^T(q_{\theta}(\hat{y}|\mathbf{x}), p(y|\mathbf{x}), c)|
]  
\nonumber 
\\
\leq 
&\mathbb{E}_{\mathbf{x}\sim p(\mathbf{x})}
[\ell^S_\omega(q_{\theta}(\hat{y}|\mathbf{x}), p(y|\mathbf{x}), c)]
] + \label{eqn:bound2}\\
&\mathbb{E}^{1/2}_{\mathbf{x}\sim p(\mathbf{x})}
[
(\ell^S_\omega(q_{\theta}(\hat{y}|\mathbf{x}), p(y|\mathbf{x}), c) -
\ell^T(q_{\theta}(\hat{y}|\mathbf{x}), p(y|\mathbf{x}), c))^2
]  \nonumber \\
&\mbox{(Jensen's Inequality)}  \nonumber
\end{align}
\normalsize
As shown in the inequality (\ref{eqn:bound1}) and (\ref{eqn:bound2}), if we use absolute/square error loss as the discrepancy function and minimize the discrepancy term (\ref{eqn:lower}) to a small value $\epsilon$/$\epsilon^2$, then we have
\begin{align}
&\resizebox{0.98\hsize}{!}{$
\mathbb{E}_{\mathbf{x}\sim p(\mathbf{x})}
[\ell^T(q_{\theta}(\hat{y}|\mathbf{x}), p(y|\mathbf{x}), c)]
\leq \mathbb{E}_{\mathbf{x}\sim p(\mathbf{x})}
[\ell^S_\omega(q_{\theta}(\hat{y}|\mathbf{x}), p(y|\mathbf{x}), c)]
 + \epsilon$}\mbox{ .}
\nonumber
\end{align}
Therefore, since the expected true task-based loss is upper bounded by the expected surrogate loss plus the discrepancy, 
we can approximately (with an ${\epsilon}$-tolerance) learn the prediction model $q_{\theta}(\hat{y}|\mathbf{x})$ w.r.t.\ the task-based loss via solving the above bilevel optimization problem.

One straightforward idea to tackle the above bilevel optimization problem is to use Lagrangian relaxation (LR), i.e., 
\begin{align}
&\min\limits_{\theta,\omega}\;
\mathbb{E}_{\mathbf{x}\sim p(\mathbf{x})}
[\ell^S_{\omega}(q_{\theta}(\hat{y}|\mathbf{x}), p(y|\mathbf{x}), c)] +
\nonumber
\\
&\resizebox{0.95\hsize}{!}{$
\quad\lambda\mathbb{E}_{\mathbf{x}\sim p(\mathbf{x})}
[D(\ell^S_\omega(q_{\theta}(\hat{y}|\mathbf{x}), p(y|\mathbf{x}), c)
||
\ell^T(q_{\theta}(\hat{y}|\mathbf{x}), p(y|\mathbf{x}), c)
)]
$}
\nonumber
\\
&\mbox{, where $\lambda$ is a non-negative weight (we set $\lambda=1$).}
\label{eqn:LR}
\end{align}
However, given the fact that $\ell^T(q_{\theta}(\hat{y}|\mathbf{x}), p(y|\mathbf{x}), c)$ is non-differentiable, we cannot directly use gradient-based method to  minimize LR (\ref{eqn:LR}) w.r.t.\ both $\theta$ and $\omega$. 
%
%
%
%
%
%
Fortunately, though the second term in the LR (\ref{eqn:LR}) is non-differentiable w.r.t.\ $\theta$, it is differentiable w.r.t.\ $\omega$ given the fact that $\ell^T(q_{\theta}(\hat{y}|\mathbf{x}), p(y|\mathbf{x}), c)$ does not involve $\omega$ and  $\ell^S_{\omega}(q_{\theta}(\hat{y}|\mathbf{x}), p(y|\mathbf{x}), c)$ is differentiable.
Therefore, instead of minimizing LR (\ref{eqn:LR}) directly using all parameters, we propose to 
separate the optimization regarding $\theta$ and $\omega$, and only minimize the first term in LR  (\ref{eqn:LR}) w.r.t. $\theta$, i.e., 
\begin{align}
&
\resizebox{0.53\hsize}{!}{$
\min\limits_{\theta}
\mathbb{E}_{\mathbf{x}\sim p(\mathbf{x})}
[\ell^S_{\omega}(q_{\theta}(\hat{y}|\mathbf{x}), p(y|\mathbf{x}), c)]
$}
\label{eqn:first}
\\
&
\resizebox{0.53\hsize}{!}{$
\min\limits_{\omega}
\mathbb{E}_{\mathbf{x}\sim p(\mathbf{x})}
[\ell^S_{\omega}(q_{\theta}(\hat{y}|\mathbf{x}), p(y|\mathbf{x}), c)]
$} + 
\label{eqn:second}
\\
&
\resizebox{0.90\hsize}{!}{$
\quad\;\;\;\mathbb{E}_{\mathbf{x}\sim p(\mathbf{x})}
[D(\ell^S_\omega(q_{\theta}(\hat{y}|\mathbf{x}), p(y|\mathbf{x}), c)
||
\ell^T(q_{\theta}(\hat{y}|\mathbf{x}), p(y|\mathbf{x}), c)
)]
$}
\nonumber
\end{align}
\begin{figure}[t]
\centering
\includegraphics[width=\columnwidth]{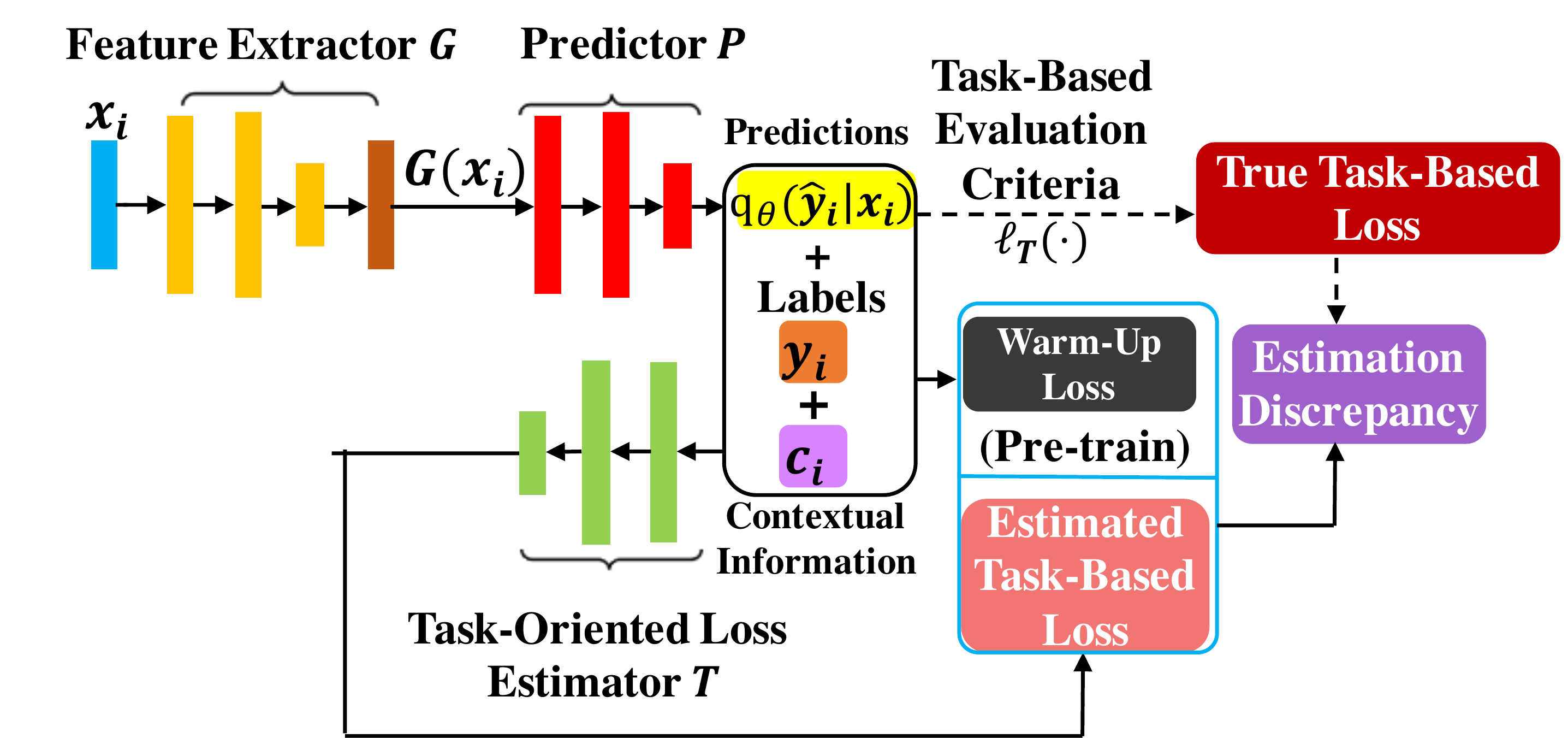}
\caption{Overview of the Task-Oriented Prediction Network.
}
\label{fig:TOPNet}
\end{figure}
\begin{algorithm}[t]
  \caption{End-to-End learning process for TOPNet 
    \label{alg:TOPNet}}
  \begin{algorithmic}[1]
    \Require{$\mathbf{x}_i$, $y_i$ and $c_i$ are raw input features, ground-truth label and corresponding contextual information sampled iid from the training set $D_{train}$.
    $ \ell^T(\cdot,\cdot, \cdot)$ is the true task-based loss function.
    $ \ell^W(\cdot,\cdot, \cdot)$ is the warm-up loss function.
    $D(\cdot||\cdot)$ is the loss discrepancy function. 
    $T,P$ and $G$ denote the task-based loss estimator, the predictor and the feature extractor respectively.
    $N_{\mbox{train}}$ is the number of training iterations.
    $N_{\mbox{pre}}$ is the number of iterations for "warm-up" pretraining.
    For ease of presentation, here we assume the batch size is 1.
    }
    \Statex
    \For{{$t \gets 1$ to $N_{\mbox{{train}}}$}}
        \State Sample a data point $(\mathbf{x}_i, y_i)$ from $D_{train}$.
        \State Make prediction $q_{\theta}(\hat{y}_i|\mathbf{x}_i) = P(G(\mathbf{x}_i))$.
        \State Invoke the true task-based criteria to compute 
        the true task-based loss $\ell^T(q_{\theta}(\hat{y}_i|\mathbf{x}_i), y_i, c_i)$.
        \State Approximate the true task-based loss using the learnable surrogate loss
        \resizebox{0.6\hsize}{!}{
        $\ell^S_{\omega_T}(q_{\theta}(\hat{y}_i|\mathbf{x}_i), y_i, c_i)=T(q_{\theta}(\hat{y}_i|\mathbf{x}_i), y_i, c_i)
        $}.
        \State Update the task-oriented estimator $T$ via $\quad\quad\quad\quad\quad$
        $
        \min\limits_{\omega_T} 
        D(\ell^S_{\omega_T}(q_{\theta}(\hat{y}_i|\mathbf{x}_i) , y_i, c_i)
        || \ell^T(q_{\theta}(\hat{y}_i|\mathbf{x}_i), y_i, c_i))
        $.
        \If{$t\leq N_{\mbox{pre}}$} 
            Update the prediction model ($P$ and $G$) using the warm-up loss:
            $\min\limits_{\theta_G,\theta_P} 
            \ell^W(q_{\theta}(\hat{y}_i|\mathbf{x}_i) , y_i, c_i)
            $.
        \Else\;Update the prediction model ($P$ and $G$) using the learned surrogate loss:
            $\min\limits_{\theta_G,\theta_P} 
            \ell^S_{\omega_T}(q_{\theta}(\hat{y}_i|\mathbf{x}_i) , y_i, c_i)
            $.
        \EndIf
    \EndFor
  \end{algorithmic}
\end{algorithm}
Intuitively, we are alternating between 
(i) optimizing the prediction model $q_\theta(\hat{y}| \mathbf{x})$ w.r.t.\ the current learned surrogate loss 
and 
(ii) minimizing the gap 
between the learned surrogate loss and the true task-based loss obtained from the current prediction model.
One can see, the learning of the prediction model and the surrogate loss depends on each other.
Thus, a bad surrogate loss would mislead the prediction model and vice versa.
For example, if the true task-based loss is a bounded loss function, then with a bad prediction model the learned surrogate loss is likely to get stuck on some insensitive area, where the loss is saturated due to the huge difference between $q_\theta(\hat{y}| \mathbf{x})$ and $p(y|\mathbf{x})$.  
Therefore, instead of starting learning the prediction model with a randomly initialized surrogate loss function, we propose to "warm-up" the prediction model $q_\theta(\hat{y}| \mathbf{x})$ with a designed warm-up loss function $\ell^W(q_{\theta}(\hat{y}|\mathbf{x}), p(y|\mathbf{x}), c)$. 
Thus, we can warm up the prediction model to be close to the ground truth so that the learning of the surrogate loss would focus more on the sensitive area and better boost the task-based performance.
In our experiments, we investigated different warm-up losses ranging from standard machine learning losses to heuristic surrogate losses.
We empirically show that the model would achieve a better performance with the "warm-up" step. 

\section{End-to-End Implementation via
Task-Oriented Prediction Network}
We instantiate the task-based learning process described above via the Task-Oriented Prediction Network (TOPNet).
As depicted in Fig.\ref{fig:TOPNet}, a feature extractor $G$ is first applied to extract meaningful features from the raw input data $\mathbf{x}_i$.
Then, a predictor network $P$ takes the extracted feature $G(\mathbf{x}_i)$ to predict the conditional distribution $P(G(\mathbf{x}_i))=q_{\theta}(\hat{y}_i|\mathbf{x}_i)$ ($\theta$ denotes the parameters in $P$ and $G$). 
Note that, in practice, we do not have access to the true distribution $p(y,\mathbf{x})$. 
Therefore we use the empirical distribution, i.e., a uniform distribution $p(y_i,\mathbf{x}_i)$ over samples in the dataset, to replace $p(y,\mathbf{x})$. 
Given the fact that the conditional distribution $p(y_i|\mathbf{x}_i)$ is indeed a Dirac Delta distribution over the value $y_i$, for ease of presentation, we use the point-wise ground truth label $y_i$ to replace the role of $p(y_i|\mathbf{x}_i)$ in the following content.
With our prediction $q_{\theta}(\hat{y}_i|\mathbf{x}_i)$, the ground truth label $y_i$ and necessary contextual information $c_i$ concerning the task, we can invoke the true task-based evaluation criteria, which potentially involve a decision-making optimization process, to generate the true task-based loss $\ell^T(q_{\theta}(\hat{y}_i|\mathbf{x}_i), y_i, c_i)$. 
Meanwhile, a task-oriented loss estimator network $T$ takes the predictions $q_{\theta}(\hat{y}_i|\mathbf{x}_i)$, the labels $y_i$, and the contextual information $c_i$, to 
approximate the true task-based loss via minimizing the discrepancy between the learned surrogate loss $\ell^S_{\omega_T}(q_{\theta}(\hat{y}_i|\mathbf{x}_i), y_i, c_i)$ ($\omega_T$ denotes the parameters in $T$) and the true task-based loss.
Finally, we can update the prediction model using the gradients obtained from the learned surrogate loss function.
As we discussed in the previous section, to facilitate the learning of both $q_\theta(\hat{y}| \mathbf{x})$ and $\ell^S_{\omega_T}(q_{\theta}(\hat{y}_i|\mathbf{x}_i) , y_i, c_i)$, we propose to warm-up the prediction model using a warm-up loss function $\ell^W(q_{\theta}(\hat{y}|\mathbf{x}), y_i, c_i)$, which could be either a standard machine learning loss or a designed heuristic loss, for the first $N_{\mbox{pre}}$ iterations.
In our experiments, we use the square error as the loss discrepancy function $D(\cdot||\cdot)$ due to its better empirical performance compared with the absolute error.
We empirically set the hyper-parameter $N_{\mbox{pre}}=|D_{train}|$ to just warm up the prediction model for one training epoch.
We summarize the implementation of the alternative minimizing process in Algorithm \ref{alg:TOPNet}.

%

\section{Experimental Results}
TOPNet is a generic learning scheme that can be used in a variety of applications with task-based criteria. In this section, we validate its performance via datasets from two real-world applications in finance. Due to business confidentiality, we are not allowed to share the datasets. 
The experiments are mainly designed to compare the benefit of using TOPNet learning scheme over standard machine learning schemes or hand-crafted heuristic surrogate loss functions. 

\paragraph{General Experimental Setup:} For all models in our experiments, the training process was done for 50 epochs, using a batch size of 1024, an Adam optimizer \cite{kingma2014adam} with a learning rate of 3e-5, and early stopping to accelerate the training process and prevent overfitting.

\subsection{Revenue Surprise Forecasting}

Revenue growth is the key indicator of the valuation and profitability of a company and it is widely used for investment decisions ~\cite{NarasimhanJegadeesh2006}, such as stock selection and portfolio management. 
Due to the long tail distribution of revenue growth, the investment communities usually predict revenue surprise which is given by revenue growth minus \textit{consensus}.
Here, \textit{consensus} is the average of the Wall street estimates of revenue growth published by stock analysts.
Despite the fact that revenues are published quarterly, daily forecasts of revenue surprise enable investors to adjust their portfolio in a granular way for return and risk analysis. 
To predict quarterly revenue surprise at the daily level before their announcement, we collect information including quarterly revenue, consensus, stock price and various of financial indicators of 
1090 US public companies ranging from Jan 1st, 2004 to June 30th, 2019. 
Each data point is associated with a 10x12-dimensional feature vector describing up-to-date sequential historical information of the corresponding company. 
The label of each data point is a real number describing the revenue surprise of the corresponding company on that specific date.
We split the whole dataset chronologically into training set (01-01-2004 to 06-30-2015, 3,267,584 data points), validation set (07-01-2015 to 06-30-2017, 465,383 data points) and test set (07-01-2017 to 06-30-2019, 421,225 data points) to validate the performance of models. 
Note that some companies only have a few data points due to their short history.
Thus, we filtered companies to make sure that all remaining  companies have enough (1,000) historical data points in the training set and end up using 902 companies in our experiments.
%
Even though we have about 4 million data points, on average each company only has about 3,600 training examples. 
Therefore, instead of learning a model for each company, we aim to use all data points to learn a company-agnostic prediction model. 
Though it is possible to build a multi-task learning framework for this specific task, it is out of the scope of this paper. 

\subsubsection{Task-based Criteria}
In this regression problem, the task-based criterion is the total reward calculated based on the Directional Accuracy (DirAcc) and the Magnitude Accuracy (MagAcc) with respect to the industry benchmark, \textit{consensus}. 
To be specific,
\small
\begin{align}
&\resizebox{0.98\hsize}{!}{$\text{DirAcc}_i = \begin{cases}
                    \alpha & \text{if } \mathrm{sign}(\tilde{\hat{y_i}})=\mathrm{sign}(\tilde{y_i})\\
                    -\beta &\text{if }  \mathrm{sign}(\tilde{\hat{y_i}})\neq\mathrm{sign}(\tilde{y_i})\\
                    \end{cases}
                    \;
\text{MagAcc}_i = \begin{cases}
                    \gamma & \text{ if } |y_i - \hat{y}_i| < 0.5|y_i|\\
                    0 & \text{otherwise}
            \end{cases}
$}\nonumber
\end{align}
\normalsize
where $\tilde{\hat{y_i}} = \hat{y_i}-\mathrm{median}(\hat{y})$,  $\tilde{y_i} = y_i-\mathrm{median}(y)$, $\hat{y_i}$ ($y_i$) denotes predicted (true) revenue surprise of a public company at a specific date, $\mathrm{sign}(\cdot)$ denotes the sign function, and $\mathrm{median(\cdot)}$ represents the median of the predicted (true) revenue surprise of data points of all the companies within the same quarter as the $i$-th data point. 
Here, we use $\mbox{DirAcc}_i$ and $\mbox{MagAcc}_i$ to denote the Directional Hit/Miss and Magnitude Hit/Miss of data point $i$, and 
$\alpha$, $\beta$ and $\gamma$ are 3 parameters denoting the reward/penalty of Directional Hit, Directional Miss, and Magnitude Hit.
In our experiments, we set $\alpha=\$5.00$, $\beta=\$6.11$ and $\gamma = \$2.22$ based on business judgement.

Intuitively, the DirAcc measures the percentage of predictions among all the companies that are more \textit{directional} accurate than the industry benchmark, which is critical for long/short investment decisions. 
The DirAcc uses the median as the anchor to adjust both our prediction and the label in order to cancel the seasonal trend within a quarter. 
The MagAcc evaluates the percentage of predictions that are significantly ($50\%$) more accurate than the industry benchmark, which is the essential input for optimizing the weight of stocks in a portfolio. 
Given $\mbox{DirAcc}_i$ and $\mbox{MagAcc}_i$, the task-based goal is to maximize the average profit the model earned from  $n$ predictions, i.e., 
$
    \frac{1}{n}
    \sum^{n}_{i=1}{\mbox{DirAcc}_i+\mbox{MagAcc}_i}.
    \label{eqn:RSF1}
$
Since algorithm \ref{alg:TOPNet} minimizes the loss function, we use the negative of equation (\ref{eqn:RSF1}) as the task-based loss in TOPNets.
\begin{figure}[t]
\centering
\includegraphics[width=7.5cm]{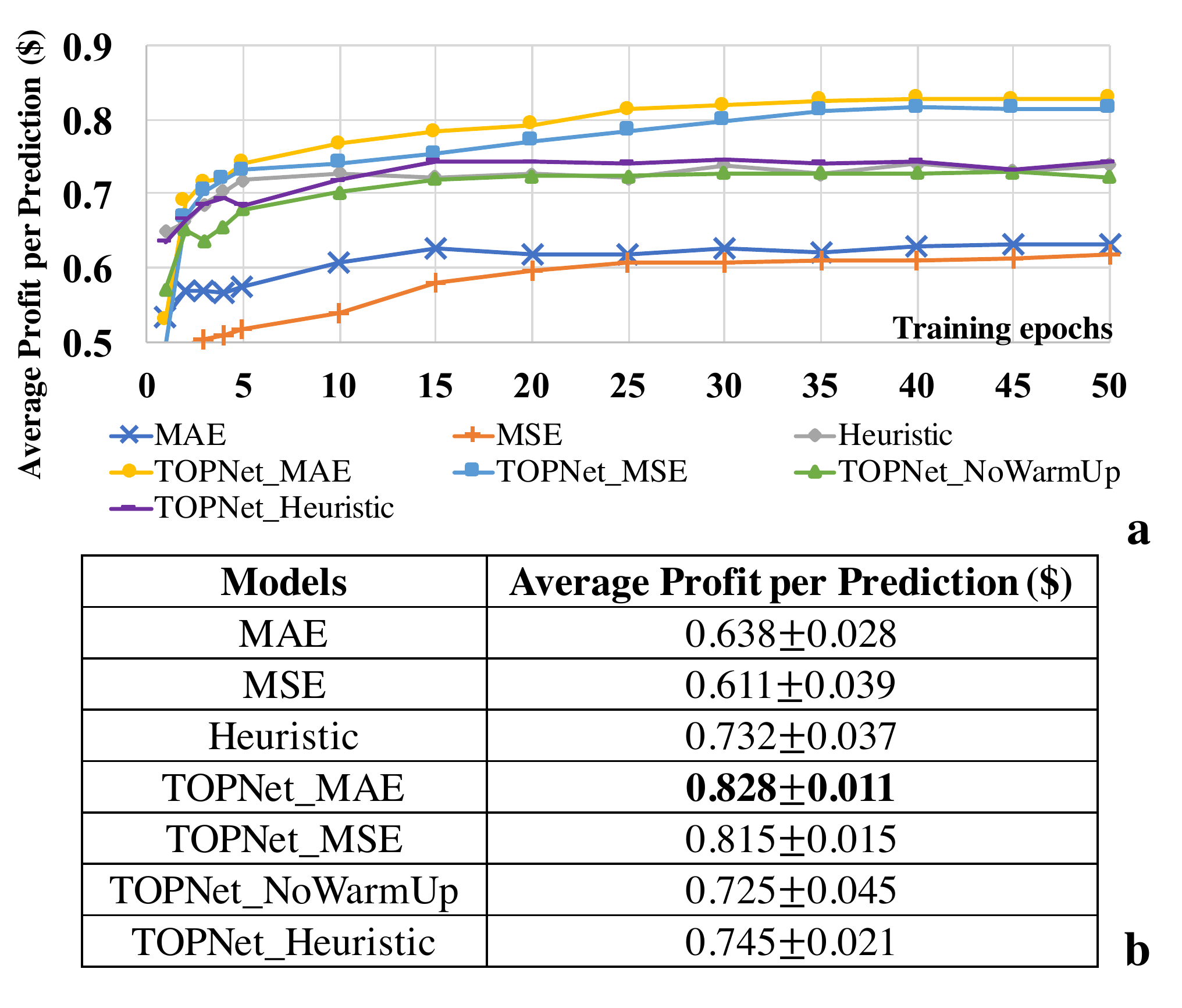}
\caption{The task-based performance of all models in the revenue surprise forecasting task. \textbf{a}. Evaluation on the validation set along the training process.
\textbf{b}. Evaluation (mean and stderr) on the test set for 15 runs of all models.
The TOPNet warmed up with MAE (TOPNet\_MAE) achieved the best performance.
}
\label{fig:RSF}
\end{figure}
\subsubsection{Benchmark Methods} 
\paragraph{(i) Models that are trained with standard machine learning loss function:}
In this regression task, we selected mean square error (MSE) loss and mean absolute error (MAE) loss as candidates of standard machine learning loss functions. 
\paragraph{(ii) Models that are trained with heuristic surrogate loss functions:}
Given the task-based criteria, we observe that a proper heuristic surrogate loss function could be designed by approximating $\mbox{DirAcc}_i$ and $\mbox{MagAcc}_i$ using $\mathrm{tanh}(\cdot)$, i.e., 
\small
\begin{align}
    &
    \resizebox{0.95\hsize}{!}{$
    \text{DirAcc}_i \approx \alpha(1 + \mathrm{sign}(\tilde{\hat{y_i}} \cdot \tilde{y_i}))/2 + 
    \beta(1 - \mathrm{sign}(\tilde{\hat{y_i}} \cdot \tilde{y_i}))/2
    $}
    \nonumber
    \\
    &
    \resizebox{0.95\hsize}{!}{$
    \approx \alpha(1 + \mathrm{tanh}(k\cdot\tilde{\hat{y_i}} \cdot \tilde{y_i}))/2 + 
    \beta(1 - \mathrm{tanh}(k\cdot\tilde{\hat{y_i}} \cdot \tilde{y_i}))/2
    $}
    \nonumber
    \\
    &\text{MagAcc}_i \approx \gamma (1+\mbox{sign}(0.5|y_i| - |y_i - \hat{y}_i|)/2)
    \nonumber
    \\
    &\approx \gamma (1 + \mathrm{tanh}(k\cdot(0.5|y_i| - |y_i - \hat{y}_i| ))/2)
    \nonumber
\end{align}
\normalsize
Here, $k$ is a scale factor and we neglect some boundary situations such as $\text{sign}(\tilde{\hat{y_i}})=\text{sign}(\tilde{y_i})=0$ and $|y_i - \hat{y}_i| = 0.5|y_i|$.  
The key idea of this approximation is to approximate $\mbox{sign}(x)$ with $\mathrm{tanh}(kx)$ since $\lim_{k \to +\infty} \mathrm{tanh}(kx) = \mathrm{sign}(x)$. 
To saturate the performance of this surrogate loss function, we exhaustively explored the best scale factor $k$ and found that it achieves the best performance with $k=100$.

\subsubsection{Experimental Setup}
We use the Long Short-Term Memory (LSTM) networks \cite{hochreiter1997long} as the feature extractors and 3-layer fully-connected neural networks as the predictors for all models in our experiments.
For a fair comparison, we explored the configuration of networks for all models to saturate their performance. 
For LSTMs and 3-layer fully-connected networks, the number of hidden units are chosen from [64, 128, 256, 512, 1024]. 
In TOPNets, the task-oriented loss estimator $T$ is a 3-layer fully-connected neural network with hidden units 1024, 512, 256. 
\subsubsection{Performance Analysis}
We did 15 runs for all models with different random seed to compute the mean and the standard error of their performance.
Since we proposed to "warm up" the predictor, we investigated the performance of TOPNets with different warm-up losses
(denoted as TOPNet\_MAE, TOPNet\_MSE, TOPNet\_Heuristic, and TOPNet\_NoWarmUp). 
As shown in Fig.\ref{fig:RSF}, TOPNets significantly outperformed the standard machine learning models trained with either MSE or MAE, boosting the average profit by about 30\%. 
TOPNets also outperformed the model trained using the hand-crafted heuristic surrogate loss function, showing the advantage of using an optimized learnable surrogate loss.
Moreover, as we expected, warming up the predictor does significantly (14\%) boost the performance compared with the TOPNet without a warm-up step (TOPNet\_NoWarmUp).
Interestingly, we observe that though the model trained with the heuristic loss alone achieved a better performance than the models trained with MSE or MAE, the heuristic loss actually made it harder to further improve the predictor with the learned surrogate loss.
The same phenomenon can also be found in the next task.

\subsection{Credit Risk Modeling}
Credit is a fundamental tool for financial transactions and many forms of economic activity. The main elements of credit risk modeling include the estimation of the probability of default and the loss given default~\cite{Doumpos2019}. 
In this study, our data includes 1.3 million personal loan applications and their payment history. Each loan is associated with an 88-dimensional feature vector and a binary label denoting whether the loan application is defaulted or not. The feature vector includes information such as the loan status (e.g., current, fully paid, default or charged off), the anonymized applicant's information (e.g., asset, debt, and FICO scores) and the loan characteristics (e.g., amount, interest rate, various cost factors of default), etc. We split the whole dataset randomly into a training set (80\%), a validation set (10\%), and a test set (10\%) to evaluate model performance.




\subsubsection{Task-based Criteria} 
The credit risk data provides information to compute the profit/loss of approving a loan application, i.e., 
\begin{align}
&\resizebox{0.95\hsize}{!}{$
    \text{Profit/Loss}= (\text{Received Principle}+
    \text{Received Interest} 
    - \text{Funded Amount}) $}
    \nonumber\\
    &\resizebox{0.5\hsize}{!}{$
    + (\text{Recovery Amount} - \text{Recovery cost}) 
    $}
    \nonumber
\end{align}
Note also that, the recovery happens only if the loan has defaulted and that if we reject a loan application, we simply earn \$0 from it.
Recall in credit risk modeling, the task-based criteria involve the prediction of the default probability $p_i$ of the $i$-th loan application as well as the probability decision threshold $p_D$ to maximize the profit after approving all loan applications with a default probability lower than $p_D$, i.e., 
\small
\begin{equation}
    \frac{1}{n}\sum^{n}_{i=1}{\mbox{Profit/Loss}_i \cdot \mbox{I}\{p_i < p_D\} + 0 \cdot \mbox{I}\{p_i \ge p_D\}}
\end{equation}
\normalsize
Here, we use $\mbox{I}\{\ \cdot\}$ to denote the indicator function.

\subsubsection{Benchmark Methods}
\paragraph{(i) Models that are trained with standard machine learning loss function:} In this 
classification task, we selected cross-entropy loss as the standard machine learning loss. 
\paragraph{(ii) Models that are trained with heuristic surrogate loss functions:}
    Given the profit/loss of approving a loan application and the predicted probability of default $p_i$, a natural surrogate loss function is, 
    $$
        (1-p_i)\cdot \mbox{profit/loss} + p_i \cdot 0,
    $$
    which measures the expected profit/loss given $p_i$.

\subsubsection{Experimental Setup}

We use 3-layer fully-connected neural networks with hidden units 1024, 512, 256 for the feature extractors $G$ of all models, and the predictors $P$ are linear layers. 
In TOPNets, the task-oriented loss estimator $T$ is a 3-layer fully-connected neural network with hidden units 1024, 512, 256. 


In this task, the evaluation criteria would optimize the decision probability threshold $p_D$ to maximize the average profit via a validation set. 
Specifically, it would sort the data points based on the predicted default probability $p_i$ and optimize the threshold $p_D$ based on the cumulative sum of the profit/loss of approving load applications with $p_i < p_D$. 
Note that, TOPNet requires point-wise task-based loss as the feedback from the task-based criteria in the training phase. 
However, computing the task-based loss involves making decisions (approve/reject), which requires the decision probability threshold $p_D$ that is supposed to be optimized on the validation set.
Noting that, the decision probability threshold $p_D$ is a relative value that depends on the predicted default probability $p_i$. 
Therefore, maintaining the order of predicted probabilities while shrinking or increasing them together does not affect the ultimate profit but leads to a different optimal threshold. 
Conversely, given a fixed decision threshold $p_D$ (e.g., 0.5), we can learn a predictor that predicts the default probability with respect to the threshold.
Thus, in the learning process of TOPNet, we used a fixed decision threshold (0.5) to make decisions and provide task-based losses in Algorithm \ref{alg:TOPNet}.
During the test, we still apply the same threshold optimization process on the predictions made by TOPNets as other models.  


\begin{table}[t]
\newcommand{\tabincell}[2]{\begin{tabular}{@{}#1@{}}#2\end{tabular}}
\centering
\resizebox{0.8\columnwidth}{!}{
\begin{tabular}{|l|c|}
\hline
\setlength\tabcolsep{2pt}
\textbf{Models} & \textbf{\makecell{Average Profit
per Loan ($\$$)}}  \\ \hline
Cross-Entropy & $618.4\pm0.3$   \\
\hline
Heuristic & $770.4\pm0.2$   \\
\hline
TOPNet\_NoWarmUp & ${770.6\pm0.2}$   \\
\hline
TOPNet\_CE & $\mathbf{784.1\pm0.2}$   \\
\hline
TOPNet\_Heuristic & $777.0\pm0.3$  \\
\hline
\end{tabular}
}
\caption{Task-based loss results (mean and stderr) of all models in the credit risk modeling task.
The TOPNet warmed-up with cross-entropy loss (TOPNet\_CE) achieved the best performance.}
\label{table:CR_cmp}
\end{table}

\subsubsection{Performance Analysis}
We did 15 runs for all models with different random seed to compute the mean and the standard error of their performance.
We evaluate the performance of TOPNets that use cross-entropy loss or heuristic loss as the warm-up loss function (denoted as TOPNet\_CE and TOPNet\_Heuristic).
We also evaluate the performance of the TOPNet without a warm-up step.
As shown in Table.\ref{table:CR_cmp}, TOPNets significantly outperformed the standard machine learning models learned with cross-entropy, boosting the average profit by $\$165.7$. 
Taking advantage of the optimized learnable surrogate loss function, the TOPNet warmed-up with cross-entropy loss further boosts the profit by $\$13.5$ per loan compared with the model trained using the heuristic loss function.
Similar to the phenomenon in 
the previous
task, the TOPNet warmed-up with the heuristic loss function performed slightly worse than the TOPNet warmed-up with cross-entropy loss. 

\section{Conclusion}
In this paper, we proposed Task-Oriented Prediction Network (TOPNet), a generic learning scheme that automatically integrates the true task-based evaluation criteria into an end-to-end learning process via a learnable surrogate loss function. 
Tested on two real-world financial prediction tasks, we demonstrate that TOPNet can significantly boost the ultimate task-based goal, outperforming both traditional modeling with standard losses and modeling with heursitic differentiable (relaxed) surrogate losses.
Future directions include exploring how to integrate 
task-based criteria that involve a strong connection 
among multiple data points.
\bibliographystyle{named}
\bibliography{main}

\end{document}